\documentclass[10pt,twocolumn,letterpaper]{article}

\usepackage{cvpr}
\usepackage{times}
\usepackage{epsfig}
\usepackage{graphicx}
\usepackage{amsmath}
\usepackage{amssymb}

\usepackage{mathtools}
\usepackage{booktabs}
\usepackage{xspace}
\usepackage{enumitem}
\usepackage{color}
\usepackage{subcaption} 
\usepackage{comment}
\usepackage{mathrsfs}

\newcommand{\full}{Teacher-Skyline}

\newcommand{\kstart}[1]{First-$#1$}
\newcommand{\kmid}[1]{Middle-$#1$}
\newcommand{\kend}[1]{Last-$#1$}
\newcommand{\sme}[1]{First-Middle-Last-$#1$}

\newcommand{\uni}[1]{Uniform-$#1$}
\newcommand{\rand}[1]{Random-$#1$}

\usepackage[pagebackref=true,breaklinks=true,letterpaper=true,colorlinks,bookmarks=false]{hyperref}

\cvprfinalcopy 

\def\cvprPaperID{6940} 

\ifcvprfinal\fi
\begin{document}

\title{Efficient Video Classification Using Fewer Frames}
\newcommand*\samethanks[1][\value{footnote}]{\footnotemark[#1]}
\author{Shweta Bhardwaj\thanks{Indian Institute of Technology Madras and Robert Bosch Centre for Data Science and AI (RBC-DSAI)}\\  {\tt\small shweta@cse.iitm.ac.in} 
        \and
        Mukundhan Srinivasan \\ \small{NVIDIA Bangalore} \\ {\tt\small msrinivasan@nvidia.com} \\
        \and
        Mitesh M. Khapra\samethanks\\  
        {\tt\small miteshk@cse.iitm.ac.in}
}


\maketitle

\begin{abstract}
Recently, there has been a lot of interest in building compact models for video classification which have a small memory footprint ($<$ 1 GB) \cite{summary-paper}. While these models are compact, they typically operate by repeated application of a small weight matrix to all the frames in a video. For example, recurrent neural network based methods compute a hidden state for \textit{every} frame of the video using a recurrent weight matrix. Similarly, \textit{cluster-and-aggregate} based methods such as \textit{NetVLAD} have a learnable clustering matrix which is used to assign soft-clusters to \textit{every} frame in the video. Since these models look at every frame in the video, the number of floating point operations (FLOPs) is still large even though the memory footprint is small. In this work, we focus on building compute-efficient video classification models which process fewer frames and hence have less number of FLOPs. Similar to memory efficient models, we use the idea of distillation albeit in a different setting. Specifically, in our case, a compute-heavy teacher which looks at all the frames in the video is used to train a compute-efficient student which looks at only a small fraction of frames in the video. This is in contrast to a typical memory efficient \textit{Teacher-Student} setting, wherein both the teacher and the student look at all the frames in the video but the student has fewer parameters. Our work thus complements the research on memory efficient video classification. We do an extensive evaluation with three types of models for video classification, \textit{viz.}, (i) recurrent models (ii) cluster-and-aggregate models and (iii) memory-efficient cluster-and-aggregate models and show that in each of these cases, a see-it-all teacher can be used to train a compute efficient see-very-little student. Overall, we show that the proposed student network can reduce the inference time by $30\%$ and the number of FLOPs by approximately $90$\% with a negligible drop in the performance. 

\end{abstract}

\section{Introduction}

Today video content has become extremely prevalent on the internet influencing all aspects of our life such as education, entertainment, communication \textit{etc.} This has led to an increasing interest in automatic video processing with the aim of identifying activities \cite{action-recog,video-beyond-short-snippet-classify}, generating textual descriptions \cite{lstm-description}, generating summaries \cite{videosumm,H-RNN}, answering questions \cite{tgif-qa} and so on. On one hand, with the availability of large-scale datasets \cite{UCF101,VideoSet-Dataset,HMDB-dataset,Youtube8m,sun-database} for various video processing tasks, it has now become possible to train increasingly complex models which have high memory and computational needs but on the other hand there is a demand for running these models on low power devices such as mobile phones and tablets with stringent constraints on latency, memory and computational cost. It is important to balance the two and design models which can learn from large amounts of data but still be computationally cheap at inference time.

In this context, the recently concluded ECCV workshop on YouTube-8M Large-Scale Video Understanding (2018) \cite{summary-paper} focused on building memory efficient models which use less than 1GB of memory. The main motivation was to discourage the use of ensemble based methods and instead focus on memory efficient single models. One of the main ideas explored by several participants \cite{paper1,paper2,paper3} in this workshop was to use knowledge distillation to build more compact student models. More specifically, they first train a teacher network which has a large number of parameters and then use this network to guide a much smaller student network which has limited memory requirements and can thus be employed at inference time. Of course, in addition to requiring less memory, such a model would also require fewer FLOPs as the size of weight matrices, hidden representations, \textit{etc.} would be smaller. However, there is scope for reducing the FLOPs further because existing models process all frames in the video which may be redundant.    

Based on the results of the ECCV workshop \cite{summary-paper}, we found that the two most popular paradigms for video classification are (i) \textit{recurrent neural network} based methods and (ii) \textit{cluster-and-aggregate} based methods. Not surprisingly, the third type of approaches based on C3D (3D convolutions) \cite{i3d} were not so popular because they are expensive in terms of their memory and compute requirements. For example, the popular I3D model \cite{i3d} is trained using 64 GPUs as mentioned in the original paper. Hence, in this paper, we focus only on the first two paradigms. We first observe that, \textit{RNN} based methods \cite{paper2,paper5,paper9} compute a hidden representation for \textit{every} frame in the video and then compute a final representation for the video based on these frame representations. Hence, even if the model is compact due to smaller weight matrix and/or hidden representations, the number of FLOPs would still be large because this computation needs to be done for every frame in the video. Similarly, \textit{cluster-and-aggregate} based methods \cite{willow,paper1,paper3,paper4,paper10} have a learnable clustering matrix which is used for assigning soft clusters to \textit{every} frame in the video. Even if the model is made compact by reducing the size of the clustering matrix and/or hidden representations, the number of FLOPs would still be large. To alleviate this problem, in this work, we focus on building models which have fewer FLOPs and are thus computationally efficient. Our work thus complements existing work on memory efficient models for video classification.

We propose to achieve 
this 
by again using the idea of distillation wherein we train a computationally expensive \textit{teacher} network which computes a representation for the video by processing all frames in the video. We then train a relatively inexpensive \textit{student} network whose objective is to process only a few frames of the video and produce a representation which is very similar to the representation computed by the teacher. This is achieved by minimizing (i) the squared error loss between the representations of the student network and the teacher network and/or (ii) by minimizing the difference between the output distributions (class probabilities) predicted by the two networks. 
Figure \ref{diagram1} illustrates this idea where the teacher sees every frame of the video but the student sees fewer frames, \textit{i.e.}, every $j$-th frame of the video. 
 At inference time, we then use the student network for classification thereby reducing  the time required for processing the video. 

We experiment with two different methods of training the \textit{Teacher}-\textit{Student} network. In the first method (which we call \textit{Serial} Training), the teacher is trained independently and then the student is trained to match the teacher with or without an appropriate regularizer to account for the classification loss. In the second method (which we call \textit{Parallel} Training), the teacher and student are trained jointly using the classification loss as well as the matching loss. This \textit{parallel} training method is similar to on-the-fly knowledge distillation from a dynamic teacher as mentioned in \cite{parallel}. We experiment with different students, \textit{viz.}, (i) a hierarchical \textit{RNN} based model (ii) \textit{NetVLAD} and (iii) \textit{NeXtVLAD} which is a memory efficient version of \textit{NetVLAD} and was the best single model in the ECCV'18 workshop. 
We experiment with the YouTube-8M dataset and show that the smaller student network reduces the inference time by upto $30\%$ while still achieving a classification performance which is very close to that of the expensive teacher network. 

\section{Related Work}
Since we focus on the task of video classification in the context of the YouTube-8M dataset \cite{Youtube8m}, we first review some recent work on video classification and then some relevant work on model compression.

\subsection{Video Classification} One of the popular datasets for video classification is the YouTube-8M dataset which contains videos having an average length of $200$ seconds. We use this dataset in all our experiments. 
The authors of this dataset proposed a simple baseline model which treats the entire video as a sequence of one-second frames and uses a Long Short-Term Memory network (LSTM) to encode this sequence. Apart from this, they also propose some simple baseline models like Deep Bag of Frames (DBoF) and Logistic Regression \cite{Youtube8m}.
Various other classification models \cite{willow,monkey-typing,temporal-models-yt8m,aggregate-frame-features,deep-models-videos} have been proposed and evaluated on this dataset (2017 version) which explore different methods of: (i) feature aggregation in videos
(temporal as well as spatial) \cite{aggregate-frame-features,willow}, (ii) capturing the interactions between labels \cite{monkey-typing} and (iii) learning new non-linear units to model the interdependencies among the activations of the network \cite{willow}. The state-of-the-art model on the 2017 version of the Youtube-8M dataset uses \textit{NetVLAD} pooling \cite{willow} to aggregate information from all the frames of a video. 

In the recently concluded competition (2018 version), many methods \cite{paper1,paper2,paper3,paper4,paper5,paper6,paper9,summary-paper} were proposed to compress the models such that they fit in 1GB of memory. As mentioned by \cite{summary-paper}, the major motivation behind this competition was to avoid the late-stage model ensemble techniques and focus mainly on single model architectures at inference time \cite{paper9,paper3, paper5}. 
One of the top performing systems in this competition was \textit{NeXtVLAD} \cite{paper3}, which modifies \textit{NetVLAD} \cite{willow} to squeeze the dimensionality of modules (embeddings). However, this model still processes all the frames of the video and hence has a large number of FLOPs. In this work, we take this compact \textit{NeXtVLAD} model and make it compute efficient by using the idea of distillation. One clear message from this workshop was that the emphasis should be on single model architectures and not ensembles. Hence, in this work, we focus on single model-based solutions. 
\subsection{Model Compression} Recently, there has been a lot of work on model compression in the context of image classification. We refer the reader to the survey paper by \cite{survey-model-compression} 
for a thorough review of the field. For brevity, here we refer to only those papers which use the idea of distillation. For example, \cite{do-deep-really-deep,know-distill,distill,obd-know-distill} use \textit{Knowledge Distillation} to learn a more compact \textit{student} network from a computationally expensive \textit{teacher} network. The key idea is to train a shallow 
student network using soft targets (or class probabilities) generated by the teacher instead of the hard targets present in the training data. 
There are several other variants of this technique such as, \cite{fitnets} extend this idea to train a student model which not only learns from the outputs of the teacher but also uses the intermediate representations learned by the teacher as additional \textit{hints}. 
This idea of \textit{Knowledge Distillation} has also been tried in the context of pruning networks for multiple object detection \cite{obd-know-distill}, speech recognition \cite{seq-teastud-speech} and reading comprehension \cite{att-distill-nlp}. 

In the context of video classification, there is some work \cite{paper1,paper6} on using \textit{Quantization} for model compression. Some work on video-based action recognition \cite{motion-optical} tries to accelerate processing in a two-stream \textit{CNN} architecture by transferring knowledge from motion modality to optical modality. In some very recent work on video classification \cite{ijcai-rl} and video captioning \cite{less-frames-eccv} the authors use a reinforcement learning agent to select which frames to process. We do not focus on the problem of frame selection but instead focus on distilling knowledge from the teacher once fewer frames have been selected for the student. While in this work we simply select frames uniformly, the same ideas can also be used on top of an RL agent which selects the best frames but we leave this as a future work.


\begin{figure*}[t]
\centering
\includegraphics[width=0.95\linewidth]{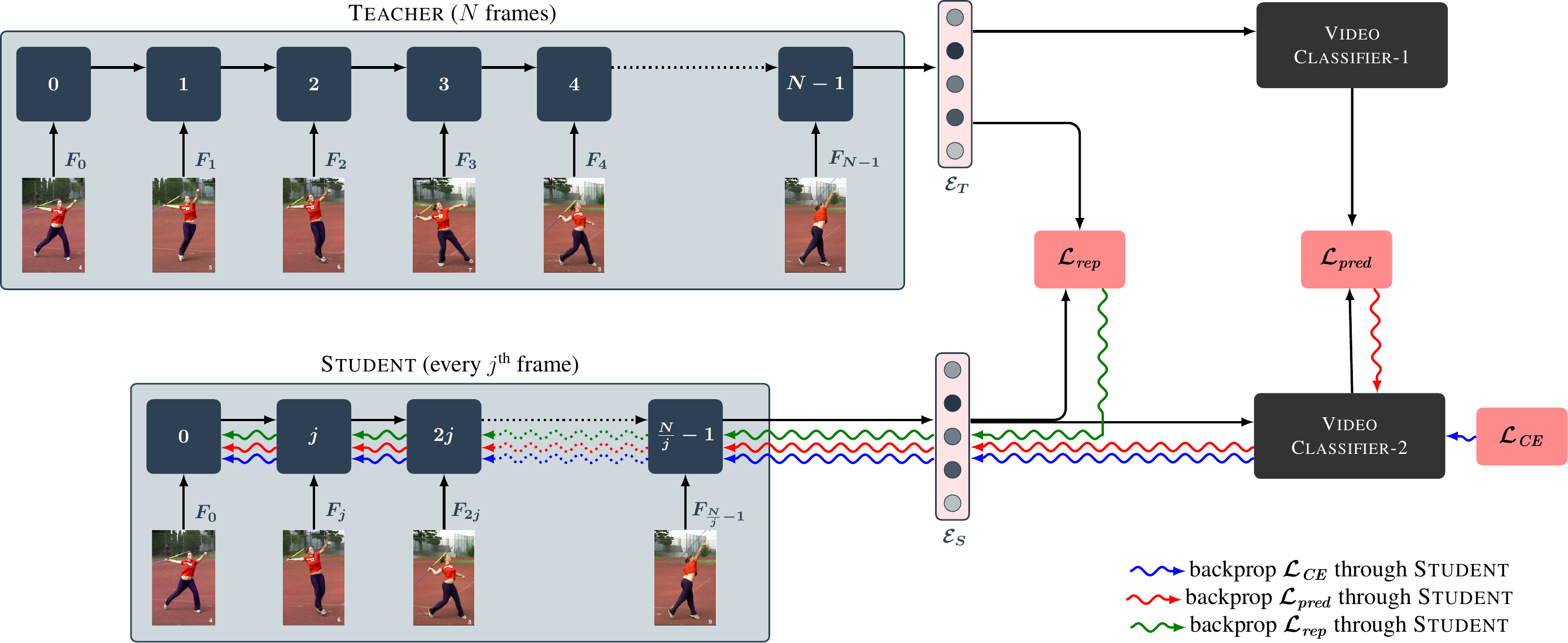}
\caption{\label{diagram1} Architecture of \textsc{Teacher-Student} network for video classification}
\end{figure*}

\section{Video Classification Models}
Given a fixed set of $m$ classes $y_1, y_2, y_3, ..., y_m \in \mathscr{Y}$ and a video $\mathbf{V}$ containing $N$ frames ($F_{0}, F_{1},\dots,F_{N-1}$), the goal of video classification is to identify all the classes to which the video belongs. In other words, for each of the $m$ classes we are interested in predicting the probability $P(y_i | \mathbf{V})$. This probability can be parameterized using a neural network $f$ which looks at all the frames in the video to predict:
\begin{align*}
P(y_i | \mathbf{V}) = f(F_{0}, F_{1},\dots,F_{N-1})
\end{align*}

Given the setup, we now briefly discuss the two state-of-the-art models that we have considered as teacher/student for our experiments. 

\subsection{Recurrent Network Based Models}
We consider the Hierarchical Recurrent Neural Network (\textit{H-RNN}) based model which assumes that each video contains a sequence of $b$ equal sized blocks. Each of these blocks in turn is a sequence of $l$ frames thereby making the entire video a sequence of sequences. In the case of the YouTube-8M dataset, these frames are one-second shots of the video and each block $b$ is a collection of $l$ such one-second frames. The model contains a lower level \textit{RNN} to encode each block (sequence of frames) and a higher level \textit{RNN} to encode the video (sequence of blocks).

\subsection{Cluster and Aggregate Based Models}
We consider the \textit{NetVLAD} model with Context Gating (CG) as proposed in \cite{willow}. This model does not treat the video as a sequence of frames but simply as a bag of frames. For every frame in this bag, it first assigns a soft cluster to the frame which results is a $M \times k$ dimensional representation for the frame where $k$ is the number of clusters considered and $M$ is the size of the initial representation of the frame (say, obtained from a CNN). Instead of using a standard clustering algorithm such as $k$-means, the authors introduce a learnable clustering matrix which is trained along with all the parameters of the network. The cluster assignments are thus a function of a parameter which is learned during training. The video representation is then computed by aggregating all the frame representations obtained after clustering. This video representation is then fed to multiple fully connected layers with Context Gating (CG) which help to model interdependencies among network activations. We also experiment with \textit{NeXtVLAD} \cite{paper3} which is a compact version of \textit{NetVLAD} wherein the $M \times k$ dimensional representation is downsampled by grouping which effectively reduces the total number of parameters in the network.

Note that all the models described above look at all the frames in the video.

\section{Proposed Approach}\label{sec-approach}
The focus of this work is to design a simpler network $g$ which looks at only a fraction of the $N$ frames at inference time while still allowing it to leverage the information from all the $N$ frames at training time. To achieve this, we propose a \textit{Teacher-Student} network as described below wherein the teacher has access to more frames than the student.

\noindent{\textbf{TEACHER:}} The teacher network can be any state-of-the-art model described above (\textit{H-RNN}, \textit{NetVLAD}, \textit{NeXtVLAD}). This teacher network looks at all the $N$ frames of video ($F_{0}, F_{1},\dots,F_{N-1}$) and computes an encoding $\mathcal{E}_{T}$ of the video, which is then fed to a simple feedforward neural network with a multi-class output layer containing a sigmoid neuron for each of the $\mathscr{Y}$ classes.
The parameters of the teacher network are learnt using  a standard multi-label classification loss $\mathcal{L}_{CE}$, which is a sum of the cross-entropy loss for each of the $\mathscr{Y}$ classes. We refer to this loss as $\mathcal{L}_{CE}$ where the subscript $CE$ stands for cross entropy between the true labels $y$ and predictions $\hat{y}$.
\begin{eqnarray}
\mathcal{L}_{CE} = - \sum_{i=1}^{|\mathscr{Y}|} y_{i} \log(\hat{y}_{i}) + (1- y_{i}) \log(1- \hat{y}_{i})\label{eq1}
\end{eqnarray}
\\
\noindent \textbf{STUDENT:}
In addition to this teacher network, we introduce a student network which only processes every $j^{th}$  frame ($F_0, F_j, F_{2j}, \dots, F_{\frac{N}{j}-1} $) of the video and computes a representation $\mathcal{E}_S$ of the video from these $\frac{N}{j}$ frames. We use only same family distillation wherein both the teacher and the student have the same architecture. For example, Figure \ref{diagram1} shows the setup when the teacher is \textit{H-RNN} and the student is also \textit{H-RNN} (please refer to the supplementary material for a similar diagram when the teacher and student are \textit{NetVLAD}). Further, the parameters of the output layer are shared between the teacher and the student. The student is trained to minimize the squared error loss between the representation computed by the student network and the representation computed by the teacher. We refer to this loss as $\mathcal{L}_{rep}$ where the subscript \textit{rep} stands for representations. 

\begin{eqnarray}
 \mathcal{L}_{rep} = || \mathcal{E}_{T} - \mathcal{E}_{S} ||^{2}\label{eq2}
\end{eqnarray}

We also try a simple variant of the model, where in addition to ensuring that the final representations $\mathcal{E}_S$ and $\mathcal{E}_T$  are similar, we also ensure that the intermediate representations ($\mathcal{I}_S$ and $\mathcal{I}_T$) of the models are similar. In particular, we ensure that the representation of the frames $j$, $2j$ and so on computed by the teacher and  student network are very similar by minimizing the squared error distance between the corresponding intermediate representations. We refer to this loss as $\mathcal{L}_{rep}^\mathcal{I}$ where the superscript $\mathcal{I}$ stands for intermediate. 
\begin{eqnarray}
 \mathcal{L}_{rep}^\mathcal{I} = \sum_{i=j,2j,..}^{\frac{N}{j}-1} ||\mathcal{I}_{T}^{i} - \mathcal{I}_{S}^{i} ||^{2}
 \label{eq3}
\end{eqnarray}
Alternately, the student can also be trained to minimize the difference between the class probabilities predicted by the teacher and the student. We refer to this loss as $\mathcal{L}_{pred}$ where the subscript $pred$ stands for \textit{predicted probabilities}. More specifically if $\mathcal{P}_{T} = \{p_T^1, p_T^2, ...., p_T^m\}$ and $\mathcal{P}_{S} = \{p_S^1, p_S^2, ...., p_S^m\}$ are the probabilities predicted for the $m$ classes by the teacher and the student respectively, then

\begin{eqnarray}
 \mathcal{L}_{pred} = d(\mathcal{P}_{T},\mathcal{P}_{S})  \label{eq4}
 \end{eqnarray}

where $d$ is any suitable distance metric such as \textit{KL} divergence or squared error loss.
 


\noindent \textbf{TRAINING:} Intuitively, it makes sense to train the teacher first and then use this trained teacher to guide the student. We refer to this as the \textit{Serial} mode of training as the student is trained after the teacher as opposed to jointly. For the sake of analysis, we use different combinations of loss function to train the student as described below:

\begin{enumerate}
\item[(a)] $\mathcal{L}_{rep}:$ Here, we operate in two stages. In the first stage, we train the student network to minimize the $\mathcal{L}_{rep}$ as defined above, \textit{i.e.}, we train the parameters of the student network to produce representations which are very similar to the teacher network. The idea is to let the student learn by only mimicking the teacher and not worry about the final classification loss. In the second stage, we then plug in the classifier trained along with the teacher (see Equation \ref{eq1}) and fine-tune all the parameters of the student and the classifier using the cross entropy loss, $\mathcal{L}_{CE}$. In practice, we found that the fine-tuning done in the second stage helps to improve the performance of the student. 

\item[(b)] $\mathcal{L}_{rep} + \mathcal{L}_{CE}$: Here, we train the student to jointly minimize the representation loss as well as the classification loss. The motivation behind this was to ensure that while mimicking the teacher, the student also keeps an eye on the final classification loss from the beginning (instead of being fine-tuned later as in the case above). 
\item[(c)] $\mathcal{L}_{pred}$: Here, we train the student to only minimize the difference between the class probabilities predicted by the teacher and the student. 

\item[(d)] $\mathcal{L}_{pred} + \mathcal{L}_{CE}$: Here, in addition to mimicking the probabilities predicted by the teacher, the student is also trained to minimize the cross entropy loss.

\item[(e)] $\mathcal{L}_{rep} + \mathcal{L}_{CE} + \mathcal{L}_{pred}$: Finally, we combine all the 3 loss functions. Figure \ref{diagram1} illustrates the process of training the student with different loss functions. 
\end{enumerate}

For the sake of completeness, we also tried an alternate mode in which we train the teacher and student in parallel such that the objective of the teacher is to minimize $\mathcal{L}_{CE}$ and the objective of the student is to minimize one of the $3$ combinations of loss functions described above. We refer to this as \textit{Parallel} training.

\begin{table*}[t]
\centering
\scalebox{0.9}{
\begin{tabular}{|l|l|c|c|c|c|c|c|c|c|c|c|}
\hline
\multicolumn{ 2}{|c|}{\textsc{Model}} & \multicolumn{ 2}{c|}{$k$=6} & \multicolumn{ 2}{c|}{$k$=10} & \multicolumn{ 2}{c|}{$k$=15} & \multicolumn{ 2}{c|}{$k$=20} & \multicolumn{ 2}{c|}{$k$=30} \\ \cline{ 3- 12}
\multicolumn{ 2}{|l|}{} & \multicolumn{1}{l|}{\textsc{GAP} } & \multicolumn{1}{l|}{mAP} & \multicolumn{1}{l|}{GAP } & \multicolumn{1}{l|}{mAP} & \multicolumn{1}{l|}{GAP } & \multicolumn{1}{l|}{mAP} & \multicolumn{1}{l|}{GAP } & \multicolumn{1}{l|}{mAP} & \multicolumn{1}{l|}{GAP } & \multicolumn{1}{l|}{mAP} \\ \hline
\multicolumn{ 2}{|l|}{\full} & \multicolumn{1}{l}{} & \multicolumn{1}{l}{} & \multicolumn{1}{l}{} & \multicolumn{1}{l}{} & \multicolumn{1}{l}{} & \multicolumn{1}{l}{} & \multicolumn{1}{l}{} & \multicolumn{1}{l|}{} & \textbf{0.811} & \textbf{0.414} \\ \hline
\multicolumn{ 2}{|l|}{\textit{Model with k frames}} & \multicolumn{7}{l}{\textsc{Baseline Methods}} & 
& \multicolumn{1}{l}{} & \multicolumn{1}{l|}{} \\ \hline
\multicolumn{ 2}{|l|}{\uni{k}} & 0.715 & 0.266 & 0.759 & 0.324 & 0.777 & 0.35 & 0.785 & 0.363 & 0.795 & 0.378 \\ 
\multicolumn{ 2}{|l|}{\rand{k}} & 0.679 & 0.246 & 0.681 & 0.254 & 0.717 & 0.268 & 0.763 & 0.329 & 0.774 & 0.339 \\ 
\multicolumn{ 2}{|l|}{\kstart{k}} & 0.478 & 0.133 & 0.539 & 0.163 & 0.595 & 0.199 & 0.632 & 0.223 & 0.676 & 0.258 \\ 
\multicolumn{ 2}{|l|}{\kmid{k}} & 0.577 & 0.178 & 0.600 & 0.198 & 0.620 & 0.214 & 0.638 & 0.229 & 0.665 & 0.25 \\ 
\multicolumn{ 2}{|l|}{\kend{k}} & 0.255 & 0.062 & 0.267 & 0.067 & 0.282 & 0.077 & 0.294 & 0.083 & 0.317 & 0.094 \\ 
\multicolumn{ 2}{|l|}{\sme{k}} & 0.640 & 0.215 & 0.671 & 0.242 & 0.680 & 0.249 & 0.698 & 0.268 & 0.721 & 0.287 \\ \hline
        \textit{Training} & \textit{Student-Loss} & \multicolumn{10}{|l|}{\textsc{\textit{Teacher-Student} Methods}} \\ \hline
        Parallel & $L_{rep}$ & 0.724 & 0.280 & 0.762 & 0.331 & 0.785 & 0.365 & 0.794 & 0.380 & \multicolumn{1}{r|}{0.803} & 0.392 \\ 
        Parallel & $L_{rep},L_{CE}$ & 0.726 & 0.285 & 0.766 & 0.334 & 0.785 & 0.362 & 0.795 & 0.381 & 0.804 & 0.396 \\ 
        Parallel & $L_{rep},L_{pred},L_{CE}$ & 0.729 & 0.292 & 0.770 & 0.337 & \textbf{0.789} & 0.371 & \multicolumn{1}{l|}{0.796} & 0.388 & \textbf{0.806} & 0.404 \\ \hline
        Serial & $L_{rep}$ & 0.727 & 0.288 & 0.768 & 0.339 & 0.786 & 0.365 & 0.795 & 0.381 & 0.802 & 0.394 \\
        Serial & $L_{pred}$ & 0.722 & 0.287 &  0.766 & 0.341 &  0.784 &  0.367 &  0.793 &  0.383&  0.798 & 0.390\\        
        Serial & $L_{rep},L_{CE}$ & 0.728 & 0.291 & 0.769 & 0.341 & 0.786 & 0.368 & 0.794 & 0.383 & 0.803 & 0.399 \\ 
        Serial & $L_{pred},L_{CE}$ &  0.724 & 0.289 & 0.763 & 0.341 & 0.785 & 0.369 & 0.795 & 0.386 & 0.799 & 0.391 \\     
        Serial & $L_{rep},L_{pred},L_{CE}$ & \textbf{0.731} & \textbf{0.297} & \textbf{0.771} & \textbf{0.349} & \textbf{0.789} & \textbf{0.375} & \textbf{0.798} & \textbf{0.390} &\textbf{ 0.806} & \textbf{0.405} \\ \hline
\end{tabular}
}
\caption{Performance comparison of proposed \textit{Teacher}-\textit{Student} models using different \textit{\textbf{Student-Loss}} variants, with their corresponding baselines using $k$ frames. Teacher-Skyline refers to the default model which process all the frames in a video.}
\label{tab-perform}
\end{table*}

\section{Experimental Setup}
In this section, we describe the dataset used for our experiments, the hyperparameters that we considered, the baseline models that we compared with and the effect of different loss functions and training methods.\\

\noindent \textbf{1. Dataset:}
The YouTube-8M dataset (2017 version) \cite{Youtube8m} contains 8
million videos with multiple classes associated with each video. The average length of a video is $200s$ and the maximum length of a video is $300s$. The authors of the dataset have provided pre-extracted audio and visual features for every video such that every second of the video is encoded as a single frame. The original dataset consists of 5,786,881 training ($70\%$), 1,652,167 validation ($20\%$) and 825,602 test examples ($10\%$). Since the authors did not release the test set, we used the original validation set as test set and report results on it. In turn, we randomly sampled 48,163 examples from the training data and used these as validation data for tuning the hyperparameters of the model. We trained our models using the remaining 5,738,718 training examples.

\noindent \textbf{2. Hyperparameters:}
For all our experiments, we used Adam Optimizer with the initial learning rate set to $0.001$  and then decreased it exponentially with $0.95$ decay rate. We used a batch size of $256$. For both the student and teacher networks we used a $2$-layered MultiLSTM Cell with cell size of $1024$ for both the layers of the hierarchical model. 
For regularization, we used dropout ($0.5$) and $\mathbf{L}_{2}$ regularization penalty of $2$ for all the parameters. We trained all the models for 5 epochs and then picked the best model-based on validation performance. We did not see any benefit of training beyond 5 epochs. For the teacher network we chose the value of $l$ (number of frames per block ) to be $20$ and for the student network, we set the value of $l$ to 5 or 3 depending on the reduced number of frames considered by the student.

In the training of \textit{NetVLAD} model, we have used the standard hyperparameter settings as mentioned in \cite{willow}. We consider $256$ clusters and $1024$ dimensional hidden layers. 
Similarly, in the case of \textit{NeXtVLAD}, we have considered the hyperparameters of the single best model as reported by \cite{paper3}. In this network, we are working with a cluster size of $128$ with hidden size as $2048$. The input is reshaped and downsampled using $8$ groups in the cluster as done in the original paper. For all these networks, we have worked with a batch size of $80$ and an initial learning rate of $0.0002$ exponentially decayed at the rate of $0.8$. Additionally, we have applied dropout of $0.5$ on the output of \textit{NeXtVLAD} layer which helps for better regularization. 


\noindent \textbf{3. Evaluation Metrics:} We used the following metrics as proposed in \cite{Youtube8m} for evaluating the performance of different models :
\begin{itemize}[leftmargin=*,noitemsep]
\item GAP (Global Average Precision): is defined as
\[GAP = \sum_{i=1}^{P} p(i) \nabla r(i)\]
where $p(i)$ is the precision at prediction $i$, $\nabla r(i)$ is the
change in recall at prediction $i$ and $P$ is the number of top predictions that we consider. Following the original YouTube-8M Kaggle competition we use the value of P as 20.
\item mAP (Mean Average Precision) : The mean average precision is computed
as the unweighted mean of all the per-class average precisions.
\end{itemize}

\noindent \textbf{4. Models Compared:} We compare our \textit{Teacher-Student} network with the following models which helps us to better contextualize our results. 
\begin{enumerate}[leftmargin=*,label=\alph*)]
\item \full: The original teacher model which processes all the frames of the video. This, in some sense, acts as the upper bound on the performance.
\item Baseline Methods: As baseline, we consider a model (\textit{H-RNN}, \textit{NetVLAD} or \textit{NeXtVLAD}) which is trained from scratch but uses only $k$ frames of the video. However, unlike the student model this model is not guided by a teacher. These $k$ frames can be (i) separated by a constant interval and are thus equally spaced (\uni{k}) or (ii) sampled randomly from the video (\rand{k}) or (iii) the first $k$ frames of the video (\kstart{k}) or (iv) the middle $k$ frames of the video (\kmid{k}) or (v) the last $k$ frames of the video (\kend{k}) or (i) the first $\frac{k}{3}$, middle $\frac{k}{3}$ and last $\frac{k}{3}$ frames of the video (\sme{k}). We report results with different values of $k$ : $6$, $10$, $15$, $20$, $30$.
\end{enumerate}

\section{Discussion And Results} 
Since we have 3 different base models (\textit{H-RNN}, \textit{NetVLAD}, \textit{NeXtVLAD}), 5 different combinations of loss functions (see section \ref{sec-approach}), 2 different training paradigms (Serial and Parallel) and 5 different baselines for each base model, the total number of experiments that we needed to run to report all these results was very large. To reduce the number of experiments we first consider only the \textit{H-RNN} model to identify the (a) best baseline (\uni{k}, \rand{k}, \kstart{k}, \kmid{k}, \kend{k}, \sme{k})  (b) best training paradigm (Serial v/s Parallel) and (c) best combination of loss function. We then run the experiments on \textit{NetVLAD} and \textit{NeXtVLAD} using only the best baseline, best training paradigm and best loss function thus identified. The results of our experiments using the \textit{H-RNN} model are summarized in Table \ref{tab-perform} to Table \ref{tab-flops} and are discussed first followed by a discussion of the results using \textit{NetVLAD} and \textit{NeXtVLAD} as summarized in Tables \ref{tab-netvlad} and \ref{tab-nextvlad}:    

\noindent\textbf{1. Comparisons of different baselines:} First, we simply compare the performance of different baselines listed in the top half of Table \ref{tab-perform}. As is evident, the Uniform-$k$ baseline which looks at equally spaced $k$ frames performs better than all the other baselines. The performance gap between Uniform-$k$ and the other baselines is even more significant when the value of $k$ is small. The main purpose of this experiment was to decide the right way of selecting frames for the student network. Based on these results, we ensured that for all our experiments, we fed equally spaced $k$ frames to the student. 

\noindent\textbf{2. Comparing Teacher-Student Network with Uniform-$k$ Baseline:} As mentioned above, the Uniform-$k$ baseline is a simple but effective way of reducing the number of frames to be processed. We observe that all the teacher-student models outperform this strong baseline. Further, in a separate experiment as reported in Table \ref{tab-data} we observe that when we reduce the number of training examples seen by the teacher and the student, then the performance of the Uniform-$k$ baseline drops and is much lower than that of the corresponding \textit{Teacher-Student} network. This suggests that the \textit{Teacher-Student} network can be even more useful when the amount of training data is limited. 

\begin{figure*}[t]
    \begin{subfigure}[b]{0.33\textwidth}
        \includegraphics[width=\textwidth,height=0.2\textheight]{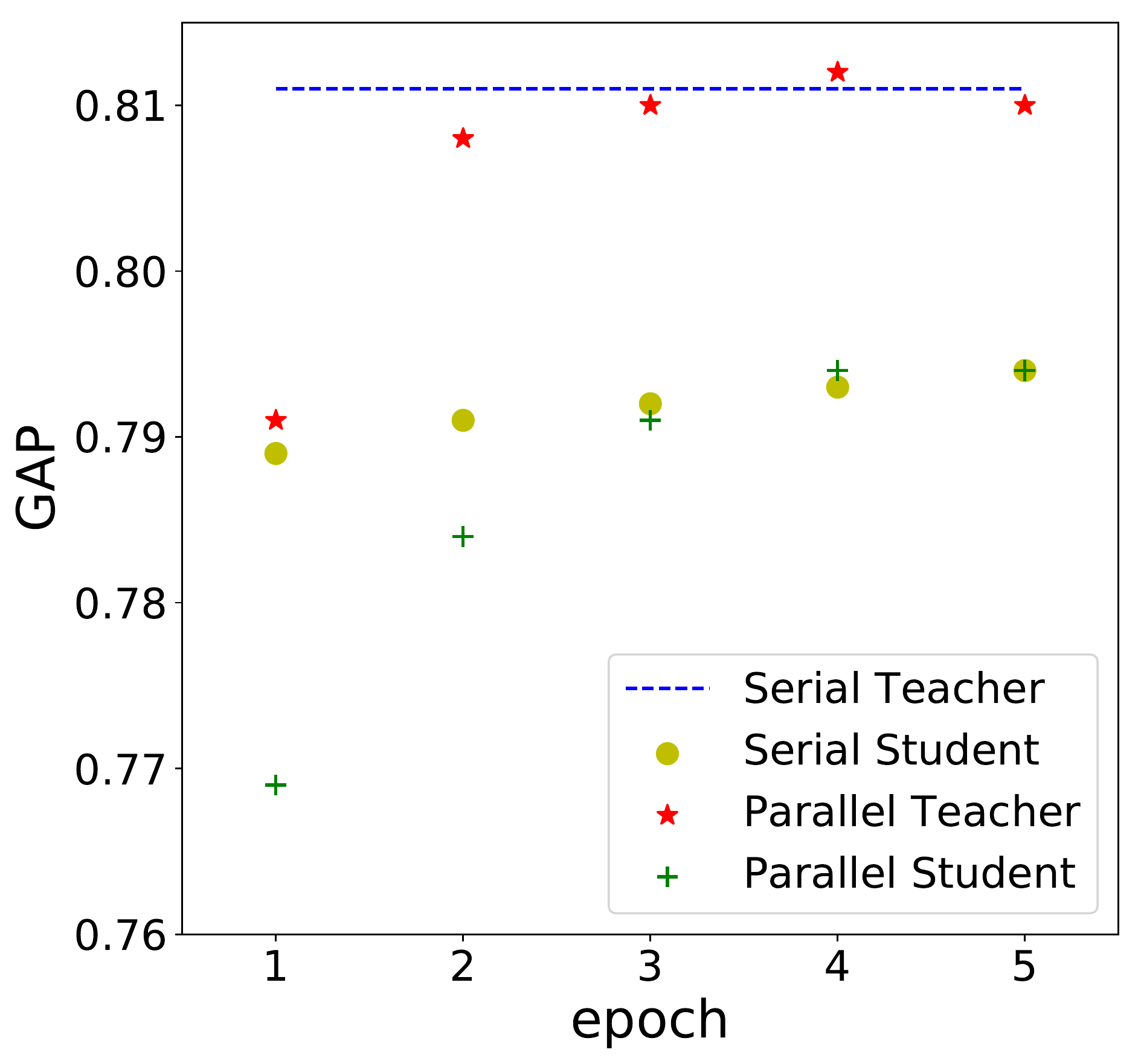}
        \caption{Training with $L_{rep}$}
        \label{plot1}
    \end{subfigure} %
    \begin{subfigure}[b]{0.33\textwidth}
      \includegraphics[width=\textwidth,height=0.2\textheight]{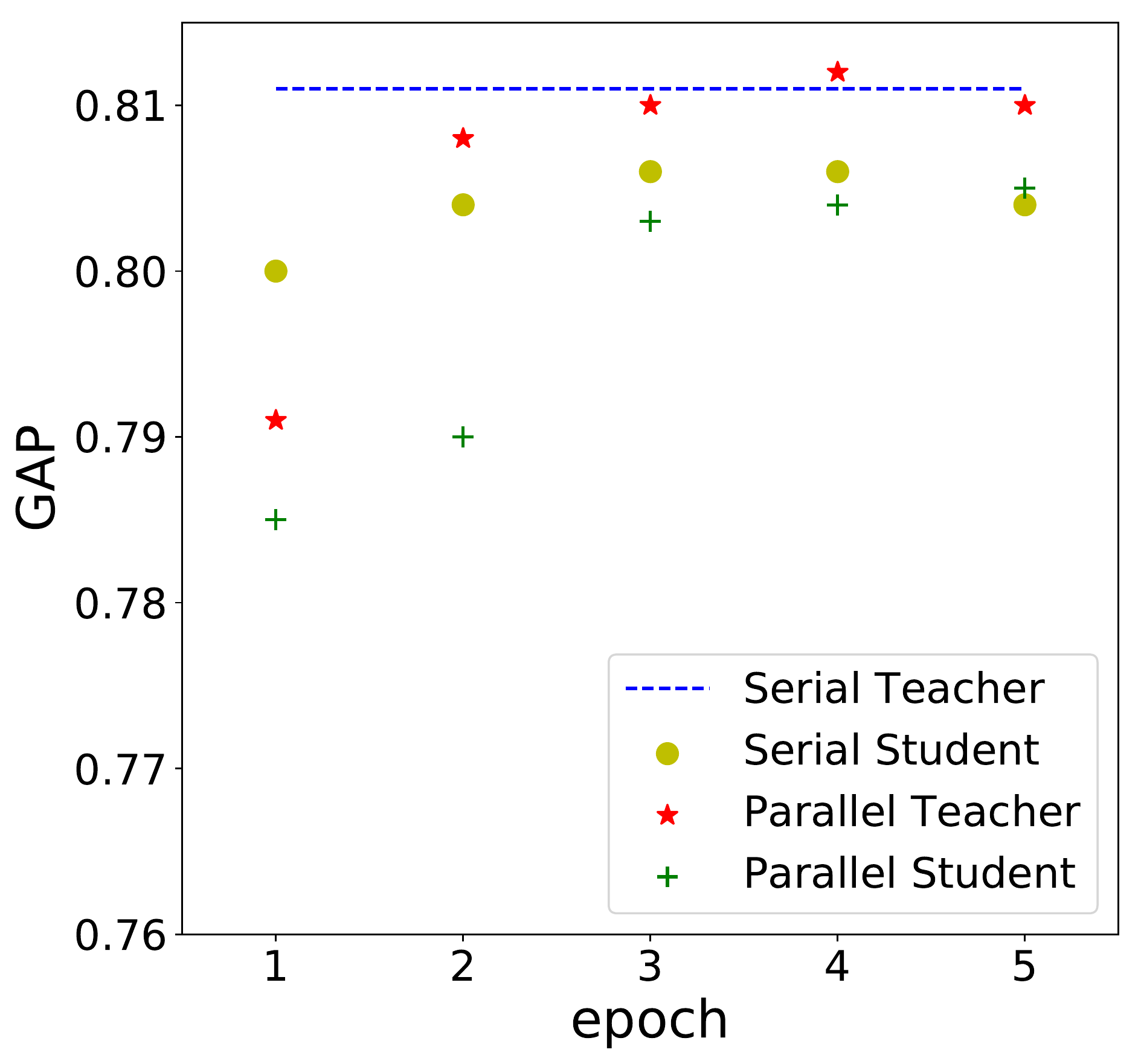}
      \caption{Training with $L_{rep}$ and $L_{CE}$}
      \label{plot2}
    \end{subfigure} %
    \begin{subfigure}[b]{0.33\textwidth}
      \includegraphics[width=\textwidth,height=0.2\textheight]{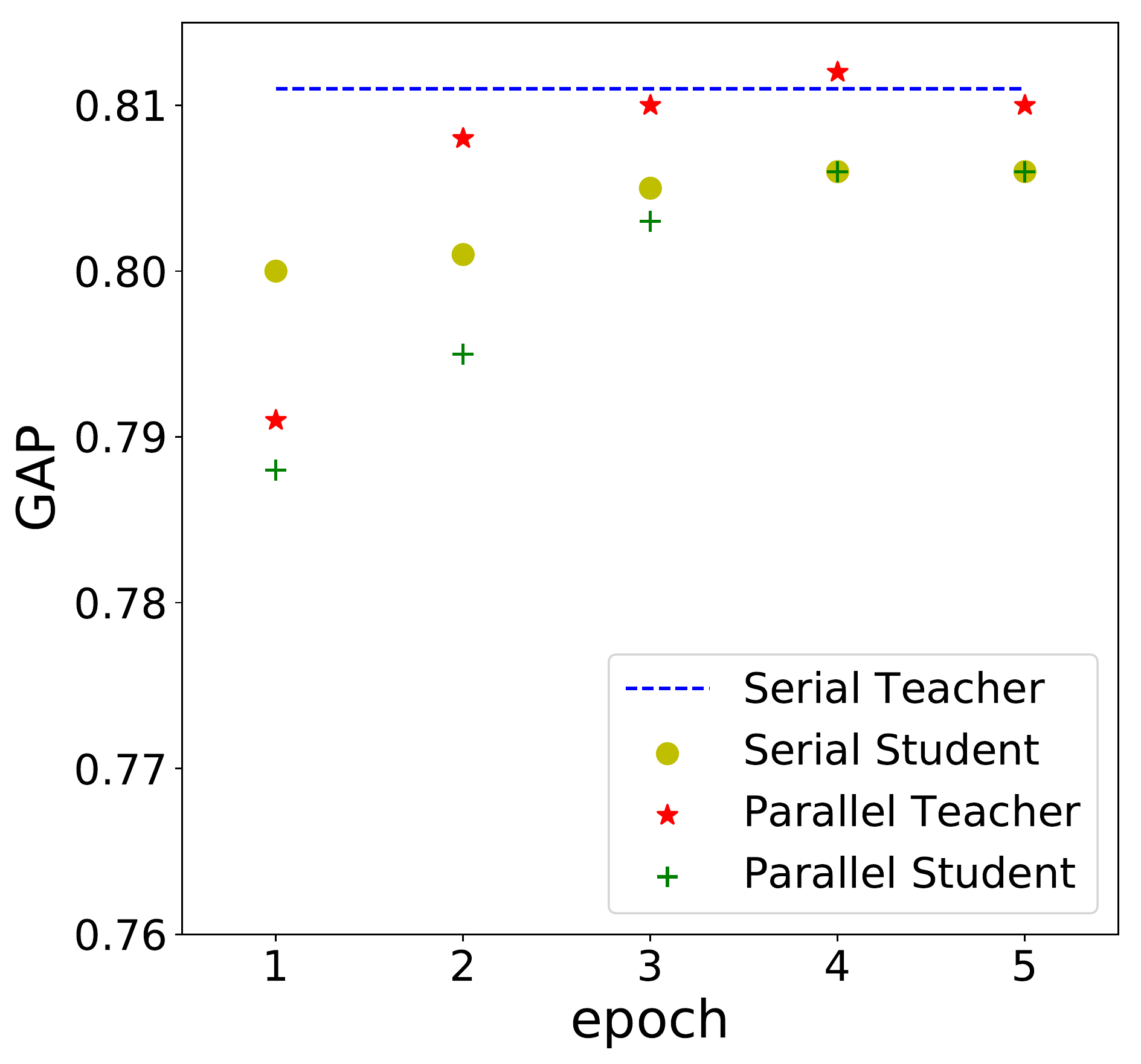}
      \caption{Training: $L_{rep}$,$L_{CE}$,$L_{pred}$}
      \label{plot3}
    \end{subfigure}
    \caption{Performance comparison (\textsc{GAP} score) of different variants of \textit{Serial} and \textit{Parallel} methods in \textit{Teacher}-\textit{Student} training.}\label{plots}
\end{figure*}
\begin{table}[t]
\centering
\begin{tabular}{l|c|ccc}
\toprule
Model & Metric & \multicolumn{3}{c}{$\%$age of training data} \\
\toprule
 & \multicolumn{1}{|l|}{} & 10$\%$ & 25$\%$ & 50$\%$ \\ 
 \toprule
Serial & GAP & 0.774 & 0.788 & 0.796 \\ 
 & mAP & 0.345 & 0.369 & 0.373 \\ 
 \midrule
Uniform & GAP & 0.718 & 0.756 & 0.776 \\ 
 & mAP & 0.220 & 0.301 & 0.349 \\ 
 \bottomrule
\end{tabular}
\caption{Effect of amount of training data on performance of Serial and Uniform models using $30$ frames}
\label{tab-data}
\end{table}
\noindent\textbf{3. Serial Versus Parallel Training of Teacher-Student:} While the best results in Table \ref{tab-perform} are obtained using \textit{Serial} training, if we compare the corresponding rows of \textit{Serial} and \textit{Parallel} training we observe that there is not much difference between the two. We found this to be surprising and investigated this further. In particular, we compared the performance of the teacher after different epochs in the \textit{Parallel} training setup with the performance of a static teacher trained independently (\textit{Serial}). We plotted this performance in Figure \ref{plots} and observed that after $3$-$4$ epochs of training, the \textit{Parallel} teacher is able to perform at par with \textit{Serial} teacher (the constant blue line). As a result, the \textit{Parallel} student now learns from this trained teacher for a few more epochs and is almost able to match the performance of the \textit{Serial} student. 
This trend is same across the different combinations of loss functions that we used.

\begin{figure}[t]
\centering
\includegraphics[width=0.9\linewidth]{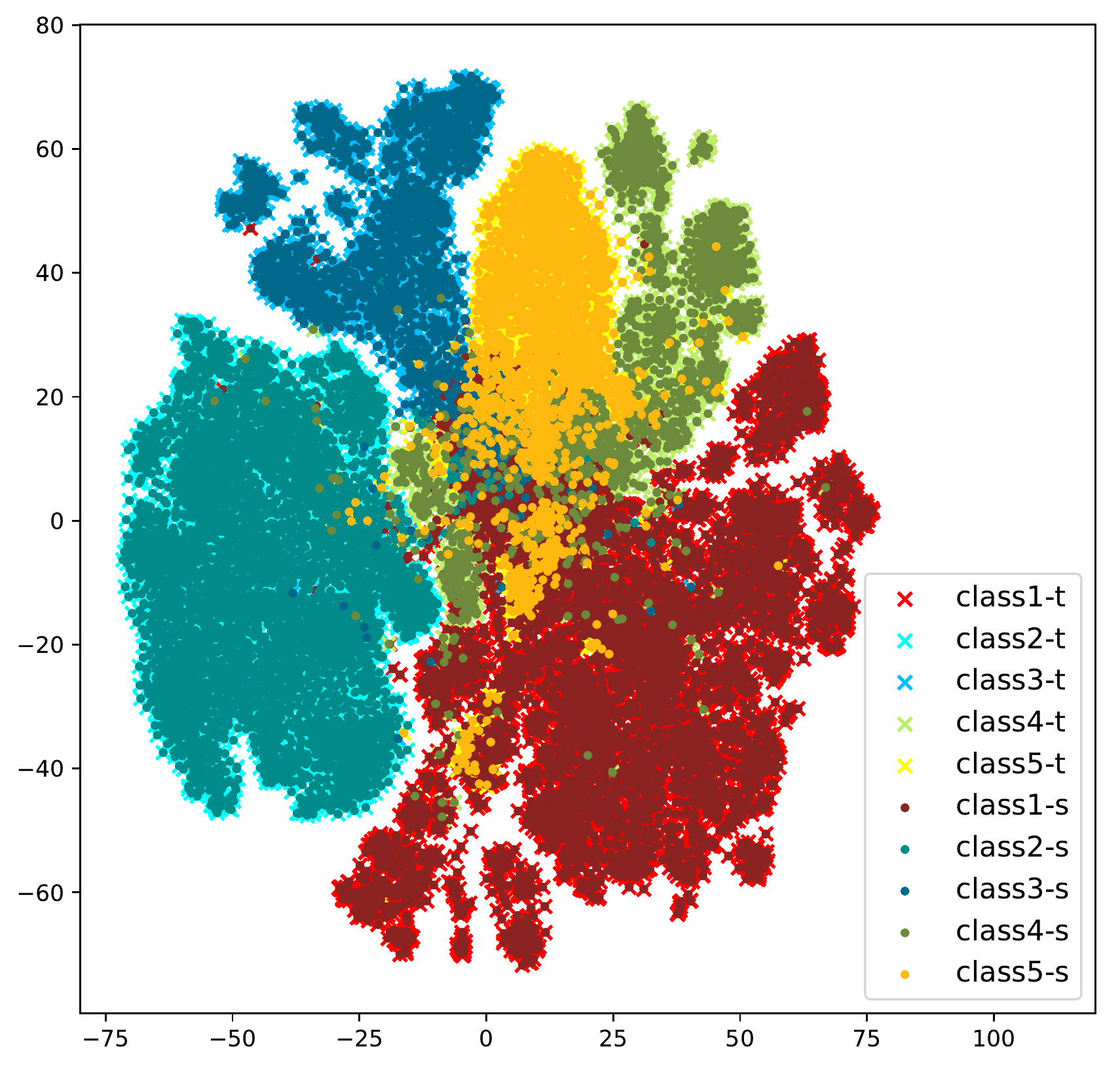}
\caption{TSNE-Embedding of teacher and student representations. Here, class $c$ refers to the cluster representation obtained corresponding to $c^{th}$ class, whereas $t$ and $s$ denote teacher and student embedding respectively.}\label{tsne}
\end{figure}

\noindent\textbf{4. Visualization of Teacher and Student Representations:} Apart from evaluating the final performance of the model in terms of mAP and GAP, we also wanted to check if the representations learned by the teacher and student are indeed similar. To do this, we chose top-$5$ classes (\textit{class1: Vehicle, class2: Concert, class3: Association football, class4: Animal, class5: Food}) in the Youtube-8M dataset and visualized the TSNE-embeddings of the representations computed by the student and the teacher for the same video (see Figure \ref{tsne}). We use the darker shade of a color to represent teacher embeddings of a video and a lighter shade of the same color to represent the student embeddings of the same video. We observe that the dark shades and the light shades of the same color almost completely overlap showing that the student and teacher representations are indeed very close to each other. This shows that introducing the $\mathscr{L}_{rep}$ indeed brings the teacher and student representations close to each other. 
\\
\begin{table}[h]
\begin{tabular}{c|c|c}
\toprule
Model  & Time (hrs.) & FLOPs (Billion)\\
\toprule
\full & 13.00 & 5.058 \\
\midrule
$k$= 10 & 7.61 & 0.167\\
$k$= 20 & 8.20 & 0.268\\
$k$= 30 & \textbf{9.11} & 0.520\\
\bottomrule
\end{tabular}
\caption{Comparison of FLOPs and evaluation time of models using $k$ frames with \textit{Skyline} model on original validation set using Tesla k80s GPU}
\label{tab-flops}
\end{table}

\noindent\textbf{5. Matching Intermediate v/s Final representations:} Intuitively, it seemed that the student should benefit more if we train it to match the intermediate representations of the teacher at different timesteps as opposed to only the final representation at the last time step. However, as reported in Table \ref{tab-intermediate}, we did not see any benefit of matching intermediate representations. 

\begin{table}[t]
\centering
\begin{tabular}{ll|cc|cc}
\toprule
\multicolumn{2}{c}{\textsc{Model}}  & \multicolumn{2}{c}{Intermediate} & \multicolumn{2}{c}{Final}\ \\
\toprule
\multicolumn{2}{c|}{ }  & \textsc{GAP} & m\textsc{AP} & \textsc{GAP} & m\textsc{AP} \\
\toprule
Parallel & $L_{rep}$ & 0.803 & 0.393 & 0.803 & 0.392\\
Parallel & $L_{rep} + L_{CE}$ & 0.803 &	0.396 & 0.804 & 0.396\\
Parallel & $L_{rep} + L_{pred}$ &	0.804 &	0.400 & 0.806 & 0.404\\
\midrule
Serial & $L_{rep}$ & 0.804 & 0.395 & 0.802 & 0.394\\
Serial & $L_{rep} + L_{CE}$ & 0.803	& 0.397 & 0.803 & 0.399\\
Serial & $L_{rep} + L_{pred}$ &	0.806 &	0.405 & 0.806 & 0.405\\
\bottomrule
\end{tabular}
\caption{Comparison of \textit{Final} and \textit{Intermediate} representation matching by \textit{Student} network using $k$=$30$ frames}
\label{tab-intermediate}
\end{table}


\begin{table}[t]
\centering
\begin{tabular}{l|cc|cc}
\toprule
  Model: \textit{NetVLAD} &  \multicolumn{ 2}{c|}{$k$=$10$} & \multicolumn{ 2}{c}{$k$=$30$} \\
\toprule
          &  mAP & GAP & mAP & GAP\\  
\midrule
  Skyline &      &     & \textbf{0.462} & \textbf{0.823} \\
  \midrule
  Uniform & 0.364 & 0.773 & 0.421 & 0.803 \\
  \midrule
  Student & 0.383 & 0.784 & 0.436 & 0.812  \\
 \bottomrule
\end{tabular}
\caption{Performance of \textit{NetVLAD} model with $k$= $10$, $30$ frames in proposed Teacher-Student Framework}
\label{tab-netvlad}
\end{table}

\begin{table}[t]
\centering
\begin{tabular}{l|cc|c}
\toprule
  Model: \textit{NeXtVLAD} &  \multicolumn{ 2}{c|}{$k$=$30$} & FLOPs \\
          &  mAP & GAP & (in Billion) \\
\midrule
  Skyline &  0.464 & 0.831 & 1.337 \\
  \midrule
  Uniform & 0.424 & 0.812 & 0.134 \\
  \midrule
  Student & 0.439 & 0.818 & 0.134 \\
 \bottomrule
\end{tabular}
\caption{Performance and FLOPs comparison in \textit{NeXtVLAD} model with $k$=$30$ frames in proposed Teacher-Student Framework}
\label{tab-nextvlad}
\end{table}

\noindent\textbf{6. Computation time of different models:} 
The main aim of this work was to ensure that the computational cost and time is minimized at inference time. The computational cost can be measured in terms of the number of FLOPs. As shown in Table \ref{tab-flops} when $k$=30, the inference time drops by $30\%$ and the number of FLOPs reduces by approximately $90\%$, but the performance of the model is not affected. In particular, as seen in Table \ref{tab-perform}, when $k=30$, the GAP and mAP drop by $0.5$-$0.9\%$ and $0.9$-$2\%$ respectively as compared to the teacher skyline. 

\noindent\textbf{7. Performance using \textit{NetVLAD} models:} 
In Table \ref{tab-netvlad} we summarize the results obtained using \textit{NetVLAD} as the base model in the \textit{Teacher-Student} network. Here the student network was trained using the best loss function ( $L_{rep},L_{pred},L_{CE}$) and the best training paradigm (\textit{Serial}) as identified from the experiments done using the \textit{H-RNN} model. Further, we consider only the \uni{k} baseline as that was the best baseline as observed in our previous experiments. Here again we observe that the student network does better than the \uni{k} baseline. 

\noindent\textbf{8. Combining with memory efficient models:} 
Lastly, we experiment with the compact \textit{NeXtVLAD} model and show that the student network performs slightly better than the \uni{k} baseline in terms of mAP but not so much in terms of GAP (note that mAP gives equal importance to all classes but GAP is influenced more by the most frequent classes in the dataset). Once again, there is a significant reduction in the number of FLOPs (approximately 89\%). 
\section{Conclusion and Future Work}
We proposed a method to reduce the computation time for video classification using the idea of distillation. Specifically, we first train a teacher network which computes a representation of the video using all the frames in the video. We then train a student network which only processes $k$ frames of the video. We use different combinations of loss functions which ensures that (i) the final representation produced by the student is the similar as that produced by the teacher and (ii) the output probability distributions produced by the student are similar to those produced by the teacher. We compare the proposed models with a strong baseline and skyline and show that the proposed model outperforms the baseline and gives a significant reduction in terms of computational time and cost when compared to the skyline.
In particular, we evaluate our model on the YouTube-8M dataset and show that the computationally less expensive student network can reduce the computation time by $30\%$ while giving an approximately similar performance as the teacher network. 

As future work, we would like to evaluate our model on other video processing tasks such as summarization, question answering and captioning. We would also like to train a student with an ensemble of teachers (preferably from different families). Lastly, we would like to train a reinforcement learning agent to first select the most favorable $k$ (or even fewer) frames in the video and use these as opposed to simply using equally spaced $k$ frames.

{\small
\bibliographystyle{ieee}
\bibliography{egbib}
}

\end{document}